\documentclass[preprint,12pt]{elsarticle}
\usepackage{multirow}
\usepackage{amsthm}
\usepackage{mathrsfs}
\usepackage{amssymb}
\usepackage[title]{appendix}
\usepackage{xcolor}
\usepackage{textcomp}
\usepackage{manyfoot}
\usepackage{booktabs}
\usepackage{algorithm}
\usepackage{algorithmicx}
\usepackage{algpseudocode}
\usepackage{listings}
\usepackage{ragged2e} 
\usepackage{subcaption}

\theoremstyle{plain} 

\theoremstyle{definition} 

\theoremstyle{remark} 

\journal{Computer Vision and Image Understanding} 

\begin{document}
\begin{frontmatter}
\title{M3D-Net: Multi-Modal 3D Facial Feature Reconstruction Network for Deepfake Detection}

\author[1]{Haotian Wu}
\author[1]{Yue Cheng}

\author[1]{Shan Bian\corref{cor1}} 
\cortext[cor1]{Corresponding author.}
\ead{bianshan@scau.edu.cn} 

\address[1]{
  {College of Mathematics and Informatics},
  {South China Agricultural University},
  {Wushan Road, Tianhe District},
  {Guangzhou},
  {510642},
  {China}
}

\begin{abstract}
With the rapid advancement of deep learning in image generation, facial forgery techniques have achieved unprecedented realism, posing serious threats to cybersecurity and information authenticity. Most existing deepfake detection approaches rely on the reconstruction of isolated facial attributes without fully exploiting the complementary nature of multi-modal feature representations. To address these challenges, this paper proposes a novel Multi-Modal 3D Facial Feature Reconstruction Network (M3D-Net) for deepfake detection. Our method leverages an end-to-end dual-stream architecture that reconstructs fine-grained facial geometry and reflectance properties from single-view RGB images via a self-supervised 3D facial reconstruction module. The network further enhances detection performance through a 3D Feature Pre-fusion Module (PFM), which adaptively adjusts multi-scale features, and a Multi-modal Fusion Module (MFM) that effectively integrates RGB and 3D-reconstructed features using attention mechanisms. Extensive experiments on multiple public datasets demonstrate that our approach achieves state-of-the-art performance in terms of detection accuracy and robustness, significantly outperforming existing methods while exhibiting strong generalization across diverse scenarios. 
\end{abstract}

\begin{keyword}
Deepfake detection \sep dual-stream network \sep 3D facial feature reconstruction \sep depth feature \sep albedo feature \sep multi-modal fusion
\end{keyword}

\end{frontmatter}

\section{Introduction}\label{sec1}
In recent years, deep learning-driven image generation technologies have advanced rapidly, propelling facial forgery techniques to unprecedented levels of realism. These technologies can now generate highly convincing facial images that are often indistinguishable from authentic ones. Broadly, deepfake techniques can be categorized into two types: one focuses on editing facial attributes or synthesizing non-existent identities, while the other manipulates facial expressions and movements in existing images or videos. While such technologies have demonstrated remarkable visual effects in fields like film and entertainment, their widespread adoption poses serious threats to cybersecurity. This growing concern has prompted research communities worldwide to actively investigate effective countermeasures against deepfake technologies.

The field of deepfake detection has witnessed the emergence of numerous advanced techniques, most of which rely on extracting visual features and identifying anomalous patterns to detect manipulated images. While these methods have demonstrated impressive performance under controlled conditions, they often encounter limitations when applied to more complex scenarios. Challenges such as data diversity, variations in image quality, and the rapid evolution of generative models continue to hinder the robustness and generalizability of existing approaches \cite{bib1}. For instance, traditional 2D feature-based detection methods tend to fail when confronted with either high-quality forgeries or images that exhibit extreme blurriness \cite{bib2}. Moreover, many current detection models exhibit a strong dependency on specific training datasets, making it difficult to generalize across different domains or datasets. These limitations highlight the need for more resilient and adaptable deepfake detection strategies.

These challenges have prompted researchers to explore more robust feature representation approaches, among which feature-reconstruction-based detection methods have garnered increasing attention. Recent reconstruction-based deepfake detection techniques attempt to enhance detection robustness by recovering spatial or geometric information from facial images to uncover latent inconsistencies indicative of manipulation. Such methods, which typically integrate facial key point extraction, deep learning networks, and 3D modeling technologies, have achieved certain research progress \cite{bib3}, yet they still face numerous challenges. First, the accuracy and realism of the reconstructed features heavily depend on the quality of the input images. In scenarios involving low-resolution faces or occlusions, the reconstruction process often lacks precision. Second, most reconstruction-based approaches focus on reconstructing a single type of feature, such as facial shape or texture—without effectively integrating multiple feature modalities. This lack of comprehensive feature fusion limits their ability to make holistic and reliable judgments about the authenticity of facial images.

To address these limitations, we propose a novel Multimodal 3D facial feature reconstruction Network for Deepfake detection, denoted as ``M3D-Net''. This is an end-to-end dual-stream network architecture that extracts facial spatial features through a 3D reconstruction module to capture subtle differences in albedo and depth features between forged and real images. It then achieves complementary advantages of the two modal features via a multi-modal fusion module with RGB features, effectively improving detection accuracy.

Our main contributions can be summarized as follows:
\begin{enumerate}
    \item  We propose a novel dual-stream network comprising a 3D facial feature reconstruction branch and an RGB branch. By reconstructing 3D facial features, the network enables multi-dimensional and in-depth feature parsing and reconstruction of facial images, which significantly enhances the performance of deepfake detection.
\item We design a novel 3D feature Pre-Fusion Module (PFM), leveraging depthwise separable convolutions and spatial kernel attention mechanism to capture multi-scale features. It effectively integrates multi-source features from the 3D reconstruction branch, yielding more discriminative feature representations.
\item We introduce a Multi-modal Fusion Module (MFM) for deep integration of RGB and 3D reconstruction features. The proposed module utilizes attention mechanisms to enable automatic modality interaction and fuses heterogeneous information, enhancing semantic understanding and robustness in complex scenarios.
\item Extensive experiments on multiple public datasets validate the effectiveness and generalization of our method. The results demonstrate that our approach achieves superior performance in terms of detection accuracy and robustness, outperforming existing state-of-the-art (SOTA) methods.
\end{enumerate}

The rest of the paper is organized as follows. Section \ref{sec2} reviews related work. Section \ref{sec3} presents the details of the proposed algorithm. Section \ref{sec4} reports experimental results. Finally, Section \ref{sec5} concludes the paper.

\section{Related works}\label{sec2}

\subsection{Deepfake detection methods}

Feature reconstruction-based deepfake detection methods primarily focus on capturing inherent feature inconsistencies in forged content that are hard to conceal. By analyzing, reconstructing, or restoring facial features at the feature level, these methods expose traces left by deepfake technologies, forming a crucial branch of current detection research. With the rapid advancement of deepfake technology, research on relevant detection methods has intensified; given the widespread dissemination of deepfake content online, detection algorithms must possess strong generalization capabilities across various manipulation types and data domains.

Li et al. \cite{bib4} introduced the ``Face X-ray'' representation, which detects whether an image is generated by blending content from two different source images. By targeting intrinsic inconsistencies in composited faces, this approach demonstrated robust generalizability across multiple facial manipulation techniques. Beyond single-modal feature analysis, Wang et al. \cite{bib5} proposed the M2TR (Multi-modal Multi-scale Transformer) network, which captures fine-grained inconsistencies across different image levels. Focusing on local artifacts and subtle manipulations, M2TR achieves high detection accuracy even for visually convincing forgeries. Guo et al. \cite{sfic} proposed SFIConv, which enhances deepfake detection by redesigning convolution operations to interactively model spatial-domain and frequency-domain features, enabling backbone networks to better capture subtle manipulation traces. Sadhya \cite{MRWC} proposed a multi-resolution wavelet convolutional network that integrates spatial features with multi-level wavelet-based frequency representations, enabling more robust deepfake detection across different data domains.

In terms of sample construction and hierarchical feature analysis, Sun et al. \cite{bib29} proposed a method to generate simulated forged training samples, termed reconstructed hybrid images. Leveraging these hybrid samples, their approach employs a Multi-scale Feature Reconstruction Network (MFRN) to perform hierarchical analysis of forgery-related artifacts (e.g., boundary anomalies and noise patterns), effectively capturing subtle differences in boundaries and textures between forged and authentic images. In the context of low-quality video detection, a specific and challenging scenario, Li et al. \cite{bib6} proposed the novel Spatial Restore Detection Framework (SRDF). Integrating super-resolution techniques and attention mechanisms, this framework uses an Enhanced Extraction Block (EEB) and a Mapping Block (MB) to recover lost spatial texture features in low-quality videos. Additionally, they introduced an improved Isolated Island Loss and a region-aware data augmentation strategy, significantly enhancing detection performance on low-resolution deepfake videos.

Self-supervised learning has also provided new insights for feature reconstruction-based detection methods. Hu et al. \cite{bib27} proposed Delocate, which learns generalizable facial part consistency features through self-supervised learning on authentic face images. The core of this approach involves training an autoencoder to reconstruct real faces by randomly masking regions of interest (ROIs); while the model excels at reconstructing authentic faces, it exhibits poor reconstruction quality on manipulated faces, thereby exposing forgery artifacts. Furthermore, although Sun et al. \cite{bib28}'s FFTG annotation method focuses primarily on annotation generation, it identifies initial regions and manipulation types using forgery masks and combines comprehensive prompting strategies to mitigate hallucination in multimodal large language models. The resulting high-quality multimodal annotations provide a valuable foundation for more accurate semantic feature reconstruction and forgery detection. Tian et al. \cite{focus} proposed FoCus, a weakly supervised framework that discovers exploitable forgery cues by generating manipulation maps from unpaired face images using only image-level labels, providing effective auxiliary supervision for multi-task deepfake detection. Baru et al. \cite{wavelet} proposed Wavelet-CLIP, a generalizable deepfake detection framework that applies wavelet-based frequency decomposition on frozen CLIP features, significantly improving cross-dataset and unseen generator detection performance. Zhang et al. \cite{hsff} proposed HSFF-Net, which enhances deepfake detection by hierarchically fusing spatial-domain and frequency-domain features to better capture subtle forgery artifacts.

In summary, despite the diverse technical paths of feature reconstruction-based deepfake detection methods, all take ``feature inconsistency'' as the core detection clue. These methods either enhance feature capture capabilities through multi-scale and multi-modal fusion, optimize model design for specific scenarios, or expand the generalizability of feature learning via self-supervised learning, providing robust technical support for deepfake detection across different scenarios.

\subsection{3D face model reconstruction methods}

3D structural understanding is pivotal in computer vision, and 3D face modeling—by capturing multi-dimensional facial attributes beyond the limitations of 2D texture parsing—enables a more comprehensive understanding of facial content, emerging as a promising direction to strengthen deepfake detection. In recent years, significant progress has been made in 3D face modeling, with its applications increasingly extending to deepfake detection tasks.

In the field of 3D face reconstruction and expression modeling from single images, Feng et al. \cite{bib7} proposed DECA (Detailed Expression Capture and Animation), a landmark method that was the first to simultaneously recover 3D facial geometry and fine-grained expression details from a single image, while supporting facial animation generation. It introduces a novel detail consistency loss to disentangle identity-specific and expression-related features. Notably, DECA requires no paired 3D supervision and demonstrates high accuracy and robustness on in-the-wild facial images, validating its potential in complex scene reconstruction and dynamic expression modeling.

Breakthroughs in unsupervised learning have further advanced 3D face modeling. Wu et al. \cite{bib8} proposed Unsup3D, an unsupervised method that leverages an autoencoder architecture to decompose a monocular image into depth, albedo, viewpoint, and illumination components. By exploiting object symmetry and learning a symmetry probability map, Unsup3D achieves effective feature disentanglement without supervision. Experiments show that it outperforms traditional 2D-supervised methods in reconstructing the 3D geometry of faces, cat faces, and even vehicles. Building on Unsup3D, Zhang et al. \cite{bib9} proposed the Learning Aggregation and Personalization (LAP) framework, which adaptively aggregates identity-inherent factors to support personalized 3D face modeling. Integrating curriculum learning strategies, a tolerant consistency loss, and an attribute refinement network, LAP better handles individual variations in unconstrained settings.

Notably, 3D face modeling technology has been directly applied to deepfake detection. Guan et al. \cite{bib31} proposed a robust face-swapping detection method based on 3D facial shape information, which takes the inconsistency between facial appearance and 3D shape as a key detection cue. Specifically, it extracts identity-related shape features using a 3D Morphable Model (3DMM) and determines the authenticity of input content by calculating the Mahalanobis distance between the extracted features and a reference template.

However, existing methods that leverage 3D facial information for forgery detection still face notable limitations. For instance, Zhang's \cite{bib10} PAD (Presentation Attack Detection) approach relies on specific hardware to capture both RGB and depth images, hindering its generalization to practical, unconstrained deepfake detection scenarios. Liu \cite{bib30} constructed an anti-spoofing dataset based on 3D facial texture and geometric information, but this method depends on a pre-constructed 3D face model repository, leading to challenges such as high data acquisition costs and limited real-time applicability. To address these issues, this paper proposes an end-to-end deepfake detection network that integrates 3D facial feature information. By deeply mining the intrinsic properties of 3D face data, the proposed framework achieves efficient and accurate identification of manipulated facial content.

In conclusion, 3D face model reconstruction methods have evolved from early general modeling to refined, personalized, and unsupervised directions. Their ability to accurately extract 3D facial structural information not only advances the field itself but also breaks through the limitations of traditional 2D vision, providing critical structural feature support for building more robust deepfake detection systems.

\section{Methodology}\label{sec3}
\subsection{Network architecture overview}\label{subsec2}

\begin{figure}[t]
\centering
\includegraphics[width=1\textwidth]{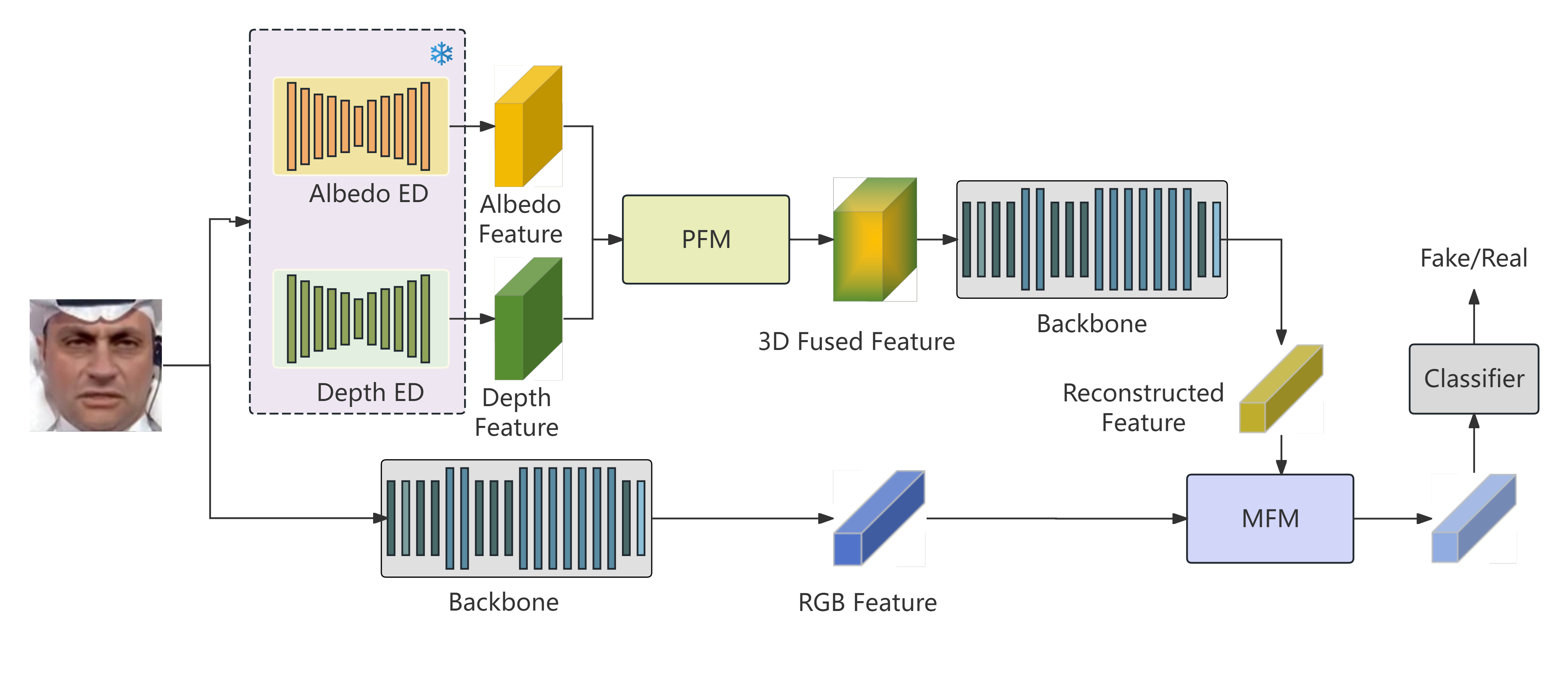}
\caption{Overall architecture of the proposed M3D-Net. The upper branch represents the 3D feature reconstruction stream, and the lower branch corresponds to the RGB stream.}\label{fig1}
\end{figure}

The architecture of the proposed M3D-Net is illustrated in Figure~\ref{fig1}. The network adopts an end-to-end dual-stream network architecture, where the input face image is fed into two branches: a 3D feature reconstruction branch and an RGB branch, each responsible for independent feature extraction. The 3D reconstruction branch incorporates two encoder-decoder (ED) structures: an Albedo Encoder-Decoder and a Depth Encoder-Decoder. These two modules are first pre-trained using a reconstruction loss function specifically designed for this task, and their parameters are subsequently frozen. Tailored for facial imagery, these encoder-decoder architectures are capable of reconstructing rich 3D spatial features of the face, which facilitates in-depth and multi-dimensional feature analysis. The features reconstructed by the Albedo and Depth modules are then integrated via the Pre-Fusion Module (PFM) designed in this study. The fused features are subsequently passed through the backbone network for further feature extraction. In parallel, the RGB branch processes the input image directly using a deep neural backbone to extract complementary features. Both branches utilize EfficientNet-B4 as the backbone network, given its proven effectiveness in feature representation. After feature extraction, the outputs of the two branches are fused by the proposed Multimodal Fusion Module (MFM), which aggregates the information into a unified feature representation for final classification and forgery detection. The following sections provide a detailed explanation of each core module in the proposed framework.
\subsection{3D facial spatial feature reconstruction module}\label{subsec2_1}
To construct a 3D facial feature extraction module for assisting deepfake detection, this work adopts Unsup3D, a framework proposed by Wu et al. \cite{bib8}, as the core component. Pretrained on real face datasets, Unsup3D enables the extraction of meaningful 3D facial features for subsequent forgery detection tasks. This method is capable of automatically estimating 3D attributes such as depth and albedo from monocular images, which are valuable for identifying structural inconsistencies commonly found in forged faces. As an unsupervised learning approach, Unsup3D does not require complex annotated data, and its strong capacity to model authentic 3D facial characteristics provides an effective means of detecting artifacts that deviate from natural facial structures in manipulated samples.

\begin{figure}[t]
\centering
\includegraphics[width=1\textwidth]{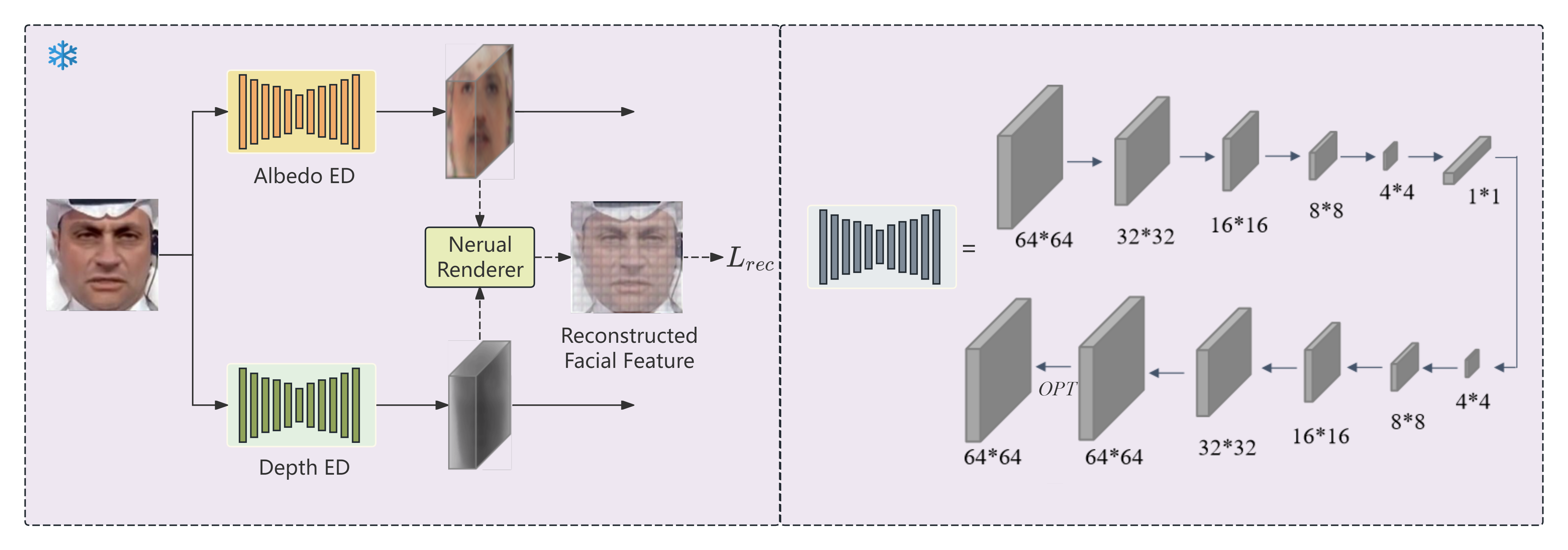}
\caption{Architecture of the 3D facial spatial feature reconstruction module.}\label{fig2}
\end{figure}

The network architecture of the 3D facial spatial feature reconstruction module is illustrated in Figure~\ref{fig2}. It primarily consists of two core encoder-decoder (ED) modules, each composed of stacked convolutional layers in the encoder and transposed convolutional (deconvolutional) layers in the decoder. In this study, we utilize two dedicated modules: the Albedo ED and the Depth ED, designed to reconstruct albedo features and depth features, respectively. The depth features represent the distance from each pixel to the observer, while the albedo features retain the intrinsic color information of the facial surface. These two sets of features are subsequently passed to the pre-fusion module for integration. This encoder-decoder architecture is specifically tailored for facial imagery and is capable of reconstructing detailed 3D spatial representations of faces, enabling comprehensive, multi-dimensional feature analysis and interpretation. Although the Albedo ED and Depth ED share an identical internal structure, they differ in the configuration of the final output layer. Specifically, the Depth ED outputs a single-channel map and employs a $tanh$ activation function to constrain the output values within the range of $(0.9, 1.1)$, suitable for representing normalized depth values. In contrast, the Albedo ED outputs a three-channel image (consistent with RGB input) with a sigmoid activation function, mapping output values to the range $(0, 1)$.

After the Albedo ED and Depth ED modules generate the albedo and depth features, these outputs are fed into a differentiable renderer, \emph{i.e.} Neural Renderer \cite{bib11}, to construct the 3D face reconstruction model. This reconstructed 3D model is directly used for computing the face reconstruction loss during the pretraining phase of this module.

\subsection{Reconstruction loss function}\label{subsec2_2}
The 3D spatial feature reconstruction module is first pretrained independently, after which its parameters are frozen during the training of the full detection network. This module is optimized using an unsupervised reconstruction loss, which consists of two components: pixel-wise reconstruction loss and perceptual loss. These two terms work together to improve model performance. The pixel-wise reconstruction loss aims to minimize the discrepancy between the reconstructed image and the original input image at the pixel level. In contrast, the perceptual loss leverages high-level semantic features extracted from a pretrained network to enforce alignment in the feature space, thereby enhancing geometric detail and mitigating the blurriness often associated with pixel-only losses. Additionally, during training, the reconstruction loss is applied to both the original and horizontally flipped versions of the input image. This design exploits the natural symmetry of human faces to further improve reconstruction accuracy.

For the reconstruction loss on the original image, an $L_1$-based loss function is employed. The mathematical formulation is as follows:
\begin{equation}
    L(\hat{I} ,I,\sigma )=-\frac{1}{\vert \Omega  \vert }  \sum_{(u,v)\in \Omega }^{}\ln \frac{1}{\sqrt{2}\sigma _{uv}  } \exp \frac{(-\sqrt{2\ell _{1,uv} } )}{\sigma _{uv} }   \label{eq1}
\end{equation}
Here, $\hat{I}$ denotes the reconstructed image, and $I$ represents the original input image. The term $\sigma$ is a confidence map indicating the model's per-pixel reconstruction certainty. The notation $\ell_{1,uv}$ refers to the $L_1$ distance between the reconstructed and original image at pixel location $(u,v)$, and $\vert \Omega \vert$ denotes the total number of pixels in the image. For the horizontally flipped image, a similar reconstruction loss is computed using the same $L_1$-based formulation, but applied between the flipped reconstructed image $\hat{I'}$ and the original (unflipped) image. This design further exploits facial symmetry to improve robustness. The final reconstruction loss is computed as a weighted sum of the two individual losses, formulated as follows:
\begin{equation}
    L_{pixel}(\phi ;I)=L(\hat{I} ,I,\sigma )+\lambda _{f} L(\hat{I} ^{'} ,I,\sigma ^{'} ) \label{eq2}
\end{equation}
Here, $\lambda_{f}$ is a weighting coefficient that balances the contributions of the two reconstruction loss terms.

In addition to the reconstruction loss, we introduce a perceptual loss to enhance the quality of geometric detail restoration. By incorporating feature constraints from a pretrained network, perceptual loss effectively captures geometric similarities that cannot be modelled by pixel-wise losses alone, thereby significantly improving the visual fidelity and structural detail of the reconstructed results \cite{bib32}. Specifically, the perceptual loss compares features extracted from the $relu3\_3$ layer of a pretrained VGG network, as illustrated in Equation~\ref{eq3}:
\begin{equation}
    L_{perc}^{(k)} (\hat{I} ',I,\sigma ^{(k)} )=-\frac{1}{\vert \Omega _{k}\vert } \sum_{(u,v)\in \Omega ^{k} }^{}\ln \left( \frac{1}{\sqrt{2\pi \sigma _{uv}^{(k)} } } \exp \left( -\frac{(\ell _{uv}^{(k)} )^{2} }{2(\sigma _{uv}^{(k)} )^{2} } \right) \right)  \label{eq3}
\end{equation}
Here, $L_{perc}^{(k)}$ denotes the perceptual loss at the $k$-th layer of the VGG network, $\hat{I'}$ represents the reconstructed image, and $I$ denotes the original input image. The term $\sigma^{(k)}$ is a confidence map corresponding to the $k$-th layer, while $\ell _{uv}^{(k)}$ represents the $L_2$ distance at spatial location $(u,v)$ in layer $k$. The symbol $\vert\Omega^k\vert$ refers to the total number of pixels in the feature map at layer $k$. The final objective function is defined as the weighted sum of the reconstruction loss and the perceptual loss, and is formulated as follows:
\begin{equation}
    L_{rec} =L_{pixel} +\lambda _{p} L_{perc}    \label{eq4}
\end{equation}
By combining the two loss functions described above, the model is jointly optimized at both the pixel level and the high-level feature representation level, enabling the reconstruction of high-quality 3D objects from single-view images under unsupervised learning conditions.

\subsection{3D Feature Pre-Fusion Module}\label{subsec2_3}
\begin{figure}[t]
\centering
\includegraphics[width=0.8\textwidth]{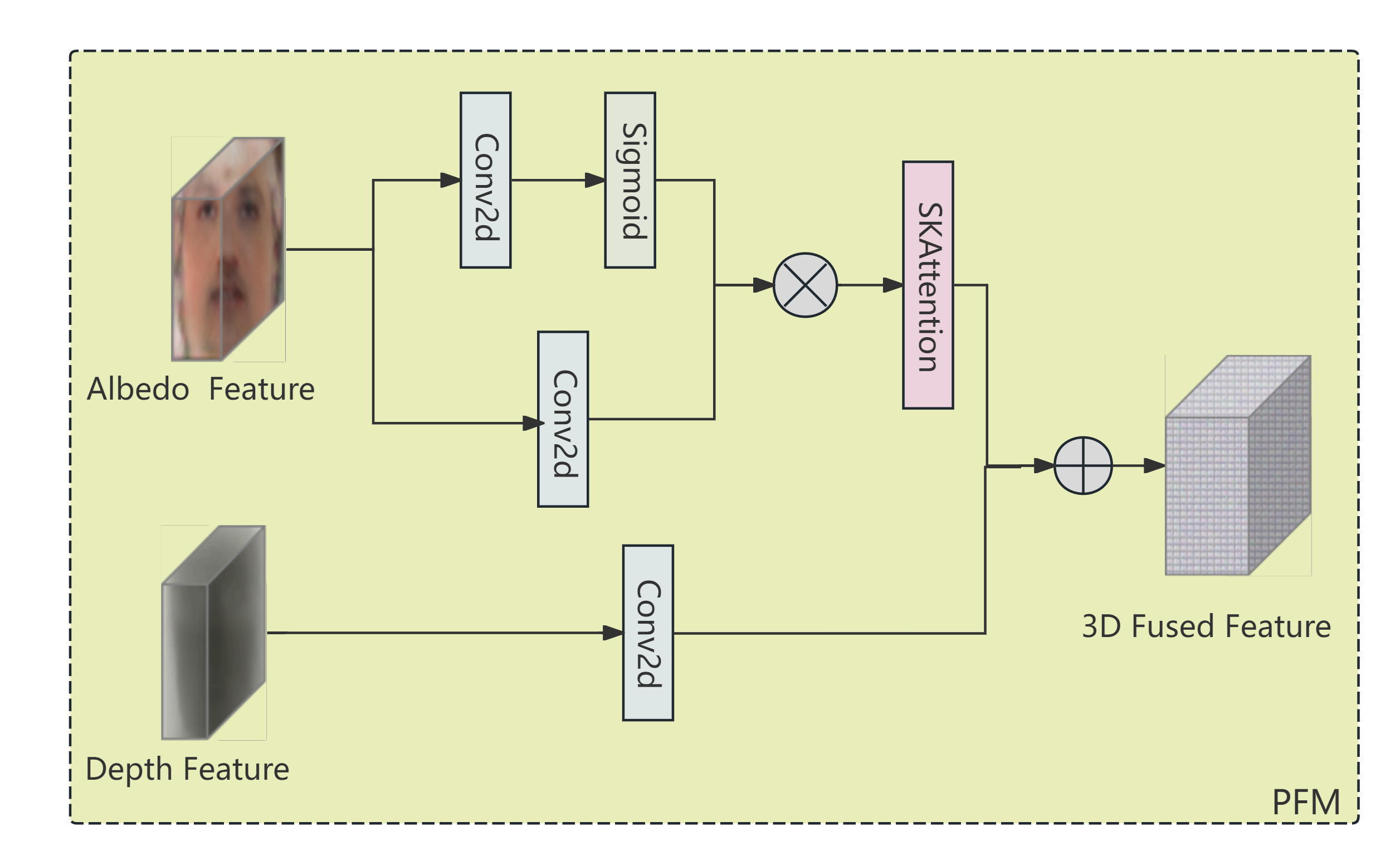}
\caption{Architecture of the Pre-Fusion Module (PFM).}\label{fig3}
\end{figure}

After obtaining the reconstructed features from the Albedo ED and Depth ED modules in the 3D reconstruction branch, we introduce a 3D Feature Pre-Fusion Module (PFM) to integrate these features. The architecture of this module is illustrated in Figure~\ref{fig3}. This module is designed to adaptively adjust and fuse albedo and depth features, allowing it to effectively cope with complex variations in facial appearance and environmental conditions. By applying feature re-weighting and deep fusion strategies, the module enhances feature expressiveness while ensuring efficient alignment and integration of multi-source information. The resulting fused representation is more discriminative, which contributes to improved detection accuracy.

The core of the module is composed of the following key components:
\begin{itemize}
    \item \textit{Depthwise Separable Convolution Layer.} Albedo image features are processed through two parallel convolutional paths, which are used to adjust the weight convolution layer and bias convolution layer respectively, so as to adaptively recalibrate the response intensity of the input feature maps. Unlike albedo image features, depth features only go through a single convolutional path. This is because depth features contain semantic information about image depth, and their differences in the channel dimension are not as significant as those of albedo image features. Therefore, a single convolutional path is sufficient to achieve effective feature extraction.
    \item  \textit{Spatial Kernel Attention Module (SKAttention)}. The SKAttention mechanism \cite{bib12} employs a multi-branch architecture with convolutional kernels of varying sizes (e.g., 1×1, 3×3, 5×5, and 7×7) to capture both local fine-grained features and broader contextual information in parallel. By integrating dimensionality reduction and expansion operations, SKAttention adaptively focuses on informative feature regions and enhances their representations. This mechanism enables the network to more effectively interpret the content and structural patterns within feature maps, improving the expressiveness and robustness of the fused representations.
\item  \textit{Feature Alignment and Fusion Layer.} This layer is responsible for aligning the re-weighted multi-spectral features with the depth features and subsequently performing a refined fusion through an attention mechanism. By jointly considering the complementary properties of both modalities, this fusion process yields more informative and robust feature representations, which enhance the network’s capacity to discriminate between real and forged facial images.
\end{itemize}

During the forward pass, the input multi-modal features are first adaptively recalibrated through depthwise separable convolutions, followed by effective alignment with the depth feature maps. The aligned features are then further refined using an attention mechanism to enable fine-grained fusion. This module not only enhances the expressiveness of the fused features but also improves the network’s adaptability to complex variations, thereby providing robust and discriminative feature representations for the deepfake detection task.

\subsection{Multimodal Feature Fusion module}\label{subsec2_4}
\begin{figure}[t]
\centering
\includegraphics[width=1.0\textwidth]{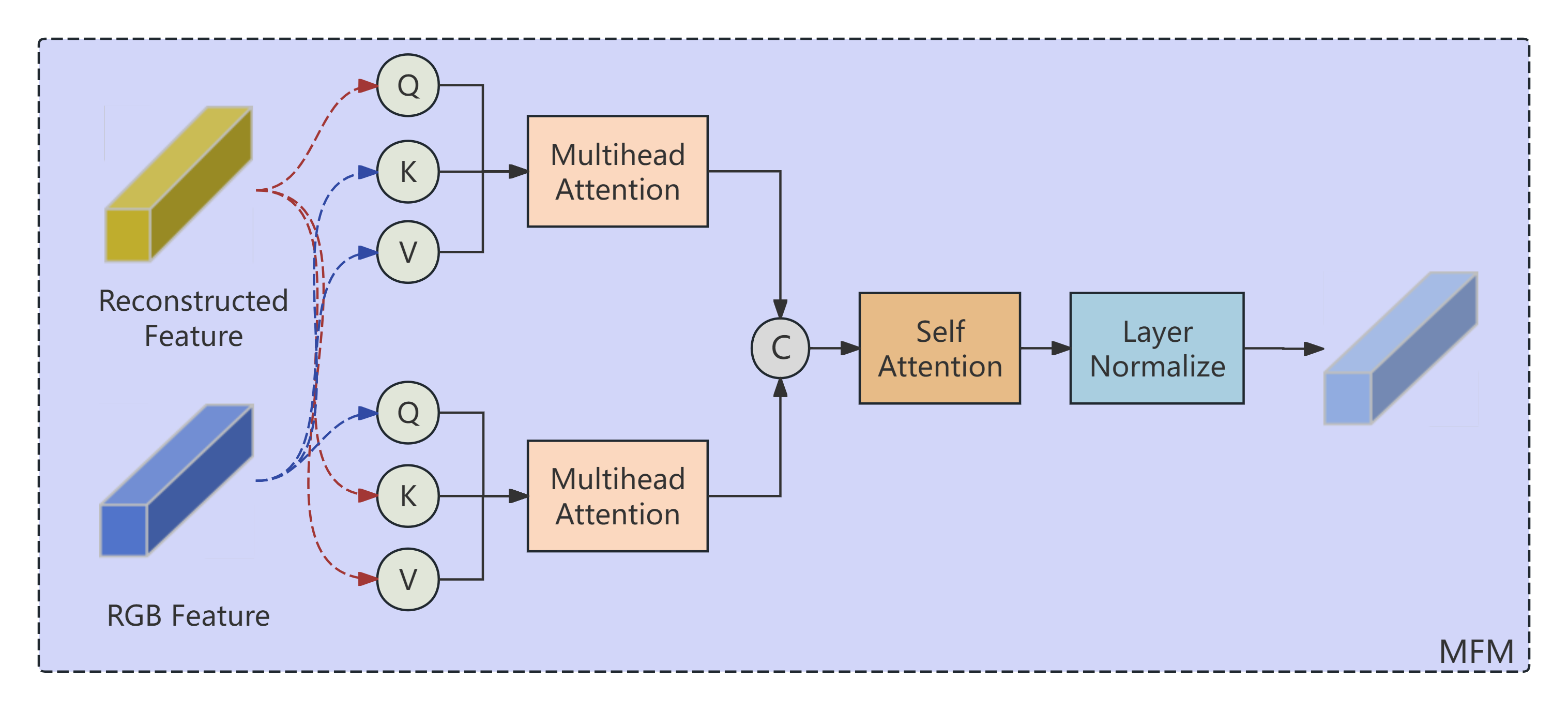}
\caption{Architecture of the Multimodal Feature Fusion Module.}\label{fig4}
\end{figure}

To integrate the high-dimensional features extracted from the two branches of the network, we propose a Multimodal Feature Fusion Module to generate the final aggregated representation for classification. The primary objective of introducing this module is to fully exploit the complementary information derived from heterogeneous modalities, thereby enhancing the model’s capacity for semantic understanding and feature representation in complex scenarios. By combining features from different modalities, the network is able to capture a more comprehensive and semantically rich representation, overcoming the limitations inherent in unimodal features. In the context of deepfake detection, RGB images provide detailed color and texture cues, while 3D facial features offer critical information about shape and depth. The fusion of these modalities facilitates a more effective differentiation between authentic and manipulated faces. Furthermore, the fusion process incorporates an attention mechanism to enable dynamic interaction and focus between modalities, thereby improving the discriminative quality and semantic expressiveness of the final feature representation. This leads to enhanced model robustness and improved detection accuracy. The detailed architecture of the Multimodal Feature Fusion Module is illustrated in Figure~\ref{fig4}.

During the feature fusion process, the module first applies linear transformation layers to project both global and local features into compatible dimensions, preparing them for the multi-head attention mechanism. Within the module, two forms of cross-attention mechanisms are constructed: one from RGB features to 3D facial features, and the other in the reverse direction—from 3D facial features to RGB features. These dual attention pathways respectively capture the guiding influence of RGB information on 3D facial representations, as well as the refinement effect of 3D geometry on RGB features, enabling more comprehensive and complementary feature interactions.

After processing through the cross-attention mechanisms, the multimodal features are concatenated along the feature dimension to form a richer and more comprehensive representation. These concatenated features are subsequently fed into a self-attention layer, which further refines and enhances their semantic information. To improve feature expressiveness, the fused features are projected via a linear layer and normalized using layer normalization, stabilizing the training process and enhancing the model’s generalization capability. The design of the Multimodal Feature Fusion Module fully exploits the strengths of attention mechanisms by enabling deep interaction between RGB and 3D facial features, facilitating efficient fusion and semantic enrichment. This architecture not only preserves the unique characteristics of each modality but also emphasizes their interrelations and complementarity, resulting in more robust and discriminative feature representations.

The multimodal features processed by the cross-attention mechanisms are concatenated along the feature dimension to form a richer and more diverse representation. This operation combines features from two distinct sources into a unified vector space, thereby integrating RGB image information and 3D facial structural information at a higher semantic level. Subsequently, the fused features are fed into a self-attention layer, which further refines and enhances their semantic content. By modeling the internal relationships within the fused features, the self-attention mechanism autonomously learns the interactions between feature components and extracts more representative and discriminative feature embeddings.

In summary, the design of the multimodal feature fusion module fully exploits the advantages of attention mechanisms. By enabling deep interactions between RGB and 3D facial features, it achieves efficient feature fusion and semantic enhancement, thereby providing strong support for improving model performance in deepfake detection tasks.

\section{Experiments}\label{sec4}
\subsection{Experimental setup}\label{subsec4_1}
To comprehensively investigate and validate the effectiveness of the proposed method, we conducted a series of controlled experiments using the DeepfakeBench framework (Yan et al.) \cite{bib14}. Although many existing studies utilize publicly available datasets for training and evaluation, slight discrepancies in data preprocessing procedures and experimental protocols often hinder fair and comparable benchmarking across methods. To address this issue, DeepfakeBench offers a unified evaluation framework with standardized metrics and experimental protocols, thereby enhancing the transparency and reproducibility of the evaluation process. In our comparative study with existing detection approaches, we adopted a dual strategy: we utilized the pre-existing experimental results provided by DeepfakeBench and also trained our model within the same platform\footnote{The code will be publicly available at https://github.com/BianShan-611/M3D-Net once the paper is accepted.}. This ensures consistent experimental conditions and guarantees the fairness and comparability of the evaluation. The detailed software and hardware configurations of the computing environment used in our experiments are presented in Table~\ref{tab1}. 

\begin{table}[h]
\centering
\caption{Experimental configuration and environment}\label{tab1}
\begin{tabular}{cc}
\toprule
Category                & Configuration        \\   \hline
Computer Type           & Desktop Computer     \\
GPU                     & Nvidia GeForce RTX 3090 \\
CPU                     & Intel Core i9-9900K  \\
Memory Size             & 64GB                 \\
Operating System        & Ubuntu 20.04 LTS     \\
Deep Learning Framework & Pytorch 1.11.0       \\
CUDA Version            & CUDA 11.3            \\
Cudnn Version           & Cudnn 8.4.0          \\
Programming Language    & Python 3.9           \\   \bottomrule
\end{tabular}
\end{table}

During the training process, a systematic strategy was employed for data preprocessing and hyperparameter configuration. Specifically, all training images were first resized to a standardized resolution of 256×256 to ensure input consistency and normalization. Each training batch consisted of 64 images (i.e., Batch Size = 64), and a series of data augmentation techniques—including random flipping, rotation, blurring, and brightness adjustment—were applied to enhance the model’s robustness and generalization capability. The Adam optimizer was used for training, with an initial learning rate set to 0.0002. To mitigate overfitting, weight decay was incorporated, and the AMSGrad variant was enabled to stabilize the optimization process. The model was trained for a total of 50 epochs, with checkpoints saved every 5 epochs to facilitate subsequent analysis and fine-tuning. For the loss function, cross-entropy loss was adopted, while the Area Under the Curve (AUC) was used as the primary evaluation metric.

To evaluate the generalization capability of the proposed framework, we conducted experiments on several widely used deepfake datasets, including FaceForensics++ (FF++) \cite{bib13}, Deepfake Detection Challenge (DFDC) \cite{bib15}, DeepfakeDetection (DFD) \cite{bib16},Deepfake Detection Challenge preview (DFDCP) \cite{bib33}, FaceShifter (Fsh) \cite{bib17}, Celeb-DF v1(CDFv1),Celeb-DF v2 (CDFv2) \cite{bib18}, and DeeperForensics-1.0 (DF-1.0) \cite{bib34}. Table~\ref{datasets} shows basic information about these datasets. FF++ contains four different face manipulation techniques: DeepFakes (DF), Face2Face (F2F), FaceSwap (FS), and NeuralTextures (NT), all applied to the same source videos. It is available in three versions corresponding to different compression levels: raw (uncompressed), lightly compressed (c23), and heavily compressed (c40). Celeb-DF v1 dataset contains 1,203 face videos of celebrities, including 408 real videos collected from YouTube with subjects of different ages, ethic groups and genders, and 795 deepfake videos synthesised from these real videos. Compared to earlier datasets, Celeb-DF v2 offers significantly improved visual quality and reduced artifacts. DeeperForensics, developed through collaboration between FAIR and MIT, encompasses diverse environments, lighting conditions, and facial expressions, ensuring high dataset variability and comprehensiveness. DFDC, released as part of a Kaggle competition, is currently the largest publicly available dataset for face forgery detection. DFDCP contains 5,244 face videos of 66 subjects with both face and voice manipulation. It was released as a preview of the full dataset of the 2020 Deepfake Detection Challenge. DFD datasets contains 3,363 face videos, covering 28 subjects, gender, and skin colour. It was created as a joint effort between two units of Google, Inc.: Google AI and JigSaw.

\begin{table}[]
\footnotesize
\centering
\caption{Overview of used datasets} \label{datasets}
\begin{tabular}{cccc}
\hline
\multirow{2}{*}{Dataset}                    & \multicolumn{2}{c}{Size}  & \multirow{2}{*}{Year} \\
                                            & Real Videos & Fake Videos &                       \\ \hline
FaceForensics++(FF++)                       & 1,000       & 4,000       & 2019                  \\
FaceShifter(Fsh)                            & 1,000       & 1,000       & 2019                  \\
Deepfake Detection Challenge preview(DFDCP) & 1,131       & 4,119       & 2019                  \\
DeepFake Detection (DFD)                    & 363         & 3,000       & 2019                  \\
Celeb-DF v1                                 & 408         & 795         & 2020                  \\
Celeb-DF v2                                 & 590         & 5,639       & 2020                  \\
DeeperForensics-1.0                         & 50,000      & 10,000      & 2020                  \\
Deepfake Detection Challenge(DFDC)          & 23,654      & 104,500     & 2020                  \\ \hline
\end{tabular}
\end{table}

To evaluate the performance of our model, we adopted AUC (the Receiver Operating Characteristic Curve), a commonly used metric in deepfake detection research. It quantifies the comprehensive performance of the model's ROC curve by assessing its ability to balance the true positive rate (TPR) and false positive rate (FPR) across varying decision thresholds. Together, these metrics provide a robust and complementary evaluation of the model’s ability to distinguish between authentic and manipulated samples. As such, they are widely adopted as standard benchmarks in the field of deepfake detection.

\subsection{Experimental results}\label{subsec4_2}
\subsubsection{Intra-dataset evaluation results}\label{subsubsec4_1}
In this section, we conduct intra-dataset comparisons between the proposed model and both baseline and state-of-the-art (SOTA) deepfake detection methods. In addition to the baseline models Xception \cite{bib22} and MesoIncep  \cite{bib23}, we include several publicly available advanced detection approaches as reference methods, including DSP-FWA  \cite{bib24}, Face X-ray, FFD \cite{bib25}, F3Net \cite{bib19}, and SRM  \cite{bib26}. The performance of all methods is evaluated using the AUC metric, as summarized in Table~\ref{tab2}. For the experiments, we used data from the four forgery techniques provided in the FaceForensics++ dataset. The comparison results are presented in Table~\ref{tab2}, with some of the reference values obtained directly from the DeepfakeBench \cite{bib14} benchmark.

\begin{table}[t]
\centering
\footnotesize
\caption{Intra-Dataset Evaluation Results (AUC) on the FaceForensics++ Dataset. The best results are shown in bold, and the second-best results are underlined.}\label{tab2}
\begin{tabular}{ccccccc}
\toprule
Method&  FF++(c23)       & DeepFake        & Face2Face       & FaceSwap        & NTextures       & Average      \\
\hline
MesoIncep               & 0.7583          & 0.8542          & 0.8087          & 0.7421          & 0.6517          & 0.763           \\
Xception                & 0.9637          & 0.9799          & 0.9785          & 0.9833          & \underline{0.9385}          & 0.9688          \\
DSP-FWA                 & 0.8765          & 0.921           & 0.9000          & 0.8843          & 0.812           & 0.8788          \\
Face X-ray              & 0.9592          & 0.9794          & \textbf{0.9872} & \underline{0.9871}          & 0.929           & 0.9684          \\
FFD                     & 0.9624          & \underline{0.9803}          & 0.9784          & 0.9853          & 0.9306          & 0.9674          \\
CORE                    & \underline{0.9638}          & 0.9787          & 0.9803          & 0.9823          & 0.9339          & 0.9678          \\
F3Net                   & 0.9635          & 0.9793          & 0.9796          & 0.9844          & 0.9354          & 0.9684          \\
SRM                     & 0.9576          & 0.9733          & 0.9696          & 0.974           & 0.9295          & 0.9608          \\
M3D-Net                 & \textbf{0.9746} & \textbf{0.9814} & \underline{0.9824}    & \textbf{0.9889} & \textbf{0.9461} & \textbf{0.9747} \\ 
\bottomrule
\end{tabular}
\end{table}

As shown in Table~\ref{tab2}, the proposed model outperforms both baseline and state-of-the-art (SOTA) methods across the overall dataset and all four forgery types. Specifically, on the FF++(c23) dataset, our method achieves an AUC score of 0.9746, representing a significant improvement over competing approaches. Our model demonstrates strong performance across a variety of manipulation scenarios, with particularly notable gains on the DeepFake, FaceSwap, and NeuralTextures subsets. However, performance on the Face2Face subset is relatively lower. This may be attributed to the unique characteristics of the Face2Face technique, which focuses on expression and motion mapping. As a result, it better preserves the nuanced facial expressions of the original subjects, making the forged content more visually consistent with real videos. This poses a greater challenge for detection models that rely on 3D feature reconstruction to distinguish between authentic and manipulated faces, thereby affecting the model's classification accuracy in this scenario. Nevertheless, our method still achieves an AUC exceeding 0.98 on the Face2Face subset, ranking second among all evaluated methods, which further demonstrates the robustness and superiority of the proposed approach.

\begin{figure}[t]
\centering
\includegraphics[width=0.8\textwidth]{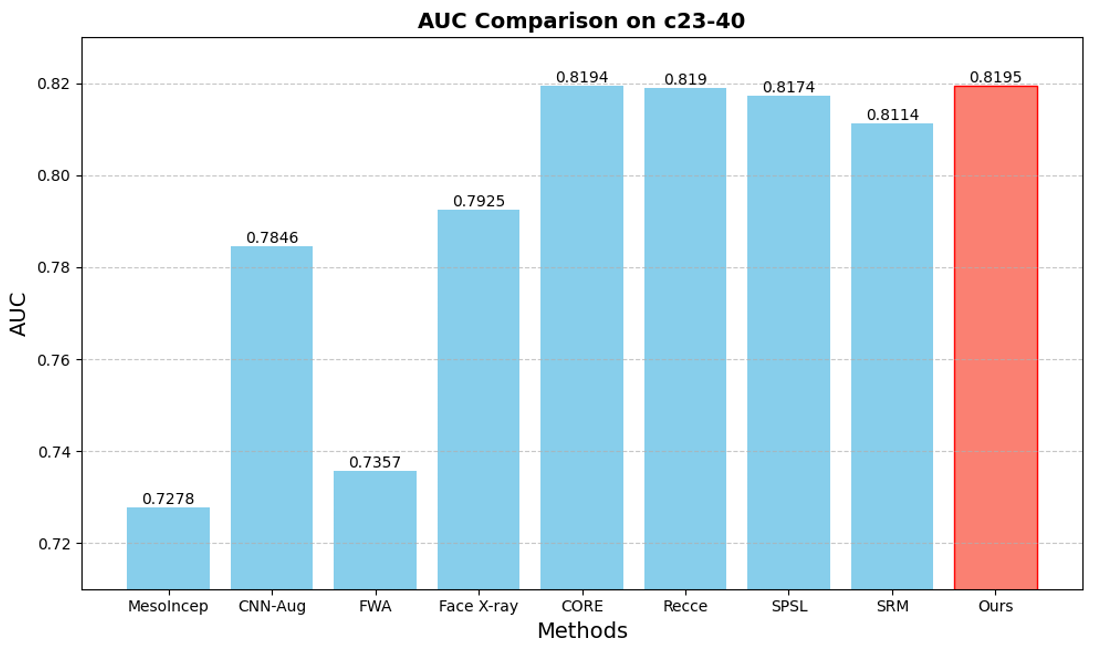}
\caption{Cross-Resolution Evaluation Results on the FF++ Dataset.}\label{fig5}
\end{figure}

To further evaluate the effectiveness of our approach, we conducted cross-resolution testing on the FF++ dataset. The results are illustrated in Figure~\ref{fig5}. Our method demonstrated outstanding performance in this setting as well, achieving an AUC score of 0.8195, outperforming all other SOTA methods. This result highlights the robustness of our model across varying input qualities: it maintains high accuracy not only on high-resolution videos but also on low-resolution ones. Such performance underscores the generalization ability of the proposed method and its adaptability to videos of different resolutions. These findings further validate the effectiveness of our approach under diverse conditions and emphasize its potential for real-world deployment.

\subsubsection{Cross-dataset evaluation results}\label{subsubsec4_2}
To assess the cross-dataset generalization capability of the proposed model, we conducted extensive experiments on several representative datasets. Specifically, the model was trained on FF++ (c23) and evaluated on Celeb-DF, DFD, DFDC, and FaceShifter (Fsh). To ensure a comprehensive and fair comparison, we included a range of both classical and recently proposed advanced detection methods as baselines, including Face X-ray \cite{bib3}, F3Net \cite{bib19}, UCF \cite{bib20}, Recce \cite{bib21}, Focus \cite{focus}, SFIC \cite{sfic}, MRWC-net \cite{MRWC}, Wavelet-net \cite{wavelet} and HSFF-net \cite{hsff}.

\begin{table}[t]
\centering
\footnotesize
\caption{Cross-Dataset Evaluation Results (AUC). The best results are shown in bold, and the second-best results are underlined.}\label{tab3}
\begin{tabular}{ccccccccc}
\hline
Method       & Year & CDFv1          & CDFv2           & DFDCP          & DFD             & DFDC            & Fsh            & DF-1.0          \\ \hline
Face X-ray   & 2020 & 0.7093          & 0.6786          & 0.6942         & 0.7655          & 0.6326          & 0.6553         & 0.5531          \\
F3Net        & 2020 & 0.7769          & 0.7352          & 0.7354         & 0.7975          & 0.7021          & 0.5914         & 0.8431          \\
FFD          & 2020 & 0.784           & 0.7435          & 0.7426         & 0.8024          & 0.7029          & 0.6056         & 0.8609          \\
SRM          & 2021 & 0.7926          & 0.7552          & 0.7408         & 0.812           & 0.6995          & 0.6014         & {\underline{ 0.8638}}    \\
Recce        & 2022 & 0.7677          & 0.7319          & 0.7419         & {\underline {0.8119}}    & 0.7133          & 0.6095         & 0.7985          \\
CORE         & 2022 & 0.7798          & 0.7428          & 0.7341         & 0.8018          & 0.7049          & 0.6032         & 0.8475          \\
UCF          & 2023 & 0.7793          & 0.7527          & 0.7594         & 0.8074          & 0.7191          & 0.6462         & 0.8241          \\
Focus        & 2024 & --              & 0.72            & \textbf{0.778} & --              & 0.669           & --             & --              \\
SFIC         & 2024 & 0.729           & 0.684           & --             & 0.764           & --              & 0.65           & --              \\
MRWC-net     & 2024 & {\underline {0.7968}}    & 0.7608          & 0.7258         & --              & --              & --             & --              \\
Wavelet-CLIP & 2025 & 0.756           & 0.759           & --             & --              & --              & \textbf{0.732} & --              \\
HSFF-net      & 2025 & --              & {\underline {0.7624}}    & --             & 0.7993          & \textbf{0.7403} & --             & --              \\
M3D-Net      &      & \textbf{0.8184} & \textbf{0.7751} & {\underline {0.7671}}   & \textbf{0.8487} & {\underline {0.7294}}    & {\underline {0.6575}}   & \textbf{0.8744} \\ 
\bottomrule
\end{tabular}
\end{table}

The results of the cross-dataset evaluation are summarized in Table~\ref{tab3}. As shown, our method outperforms existing state-of-the-art (SOTA) approaches on the CDFv1, CDFv2, DF-1.0, and DFD datasets. However, the results also reveal that our model performs slightly below some advanced methods on the DFDCP and DFDC datasets. This performance gap may be attributed to limitations of the employed 3D face reconstruction model. Specifically, Unsup3D excels at handling images with strong symmetry—an advantage rooted in its architectural design—but may struggle under cross-dataset conditions involving complex head poses, such as extreme facial tilts or occlusions. These challenges can degrade reconstruction quality, thereby affecting detection performance. It can be observed that our method achieves slightly inferior performance on the Fsh dataset compared with other datasets. We attribute this discrepancy to a mismatch between the local symmetry pirors our model relies on and the intrinisic characteristics of Fsh-generated data. The Fsh method is known for high-quality identity swapping and near-perfect illumination and texture blending, which can result in forged faces with highly realistic single -frame 3D geometric structures. As a consequence, the discriminative capability of our model is substantially weakened on this datasets. Nevertheless, our model still demonstrates strong generalization ability.
\begin{table}[t]
\centering
\footnotesize
\caption{Ablation study on Effectiveness of Backbone Network and Pre-fusion Module}\label{BP}
\begin{tabular}{cccccccc}
\hline
\multirow{2}{*}{Variation} & \multicolumn{2}{c}{Backbone} & \multirow{2}{*}{PFM} & \multicolumn{4}{c}{\textbf{}}                                         \\
                           & Xception    & EfficientNet   &                      & FF++            & CDFv2           & DFDC            & Fsh             \\ \hline
I                          & \checkmark           &                & \checkmark                    & 0.9109          & 0.7373          & 0.6466          & 0.5852          \\
II                          &             & \checkmark              &                      & 0.9735          & 0.7621          & 0.7003          & 0.6089          \\
III                          &             & \checkmark              & \checkmark                    & \textbf{0.9746} & \textbf{0.7751} & \textbf{0.7294} & \textbf{0.6575} \\ \hline
\end{tabular}
\end{table}
\subsubsection{Ablation Study}\label{subsubsec4_3}
To evaluate the contribution of each component to the overall model performance, we conducted detailed ablation studies and tested the AUC scores of different configurations across multiple datasets. 

\textit{Effectiveness of the Backbone Network}: We compared two commonly used backbone architectures: Xception and EfficientNet. As shown in Table~\ref{BP} by the comparison between Variants I and III, the model utilizing EfficientNet consistently outperformed the Xception-based counterpart across all evaluated datasets. This indicates that EfficientNet is better suited for our task, providing stronger feature representation capabilities.

\textit{Effectiveness of the Pre-Fusion Module}:We investigate the impact of pre-fusion module. As reported in Table~\ref{BP} by comparing Variants II and III, it is evident that incorporating the pre-fusion module leads to performance improvements across all evaluated datasets. For instance, on the FaceShifter dataset, the AUC increased from 0.6089 to 0.6575. This demonstrates that the pre-fusion module facilitates the integration and complementation of 3D reconstruction features, thereby enhancing the accuracy of subsequent detection.

\textit{Impact of the Attention Mechanism}: In this section, we investigate the effectiveness of our attention mechanism in MFM. The comparison results are presented in Table~\ref{taba}, reveals that incorporating the attention mechanism within the detection network generally leads to improved AUC scores across various datasets, highlighting the critical role of the proposed attention module in multimodal feature fusion.

Furthermore, Table~\ref{attn} investigate the effect of varying the number of attention heads on model performance. We evaluated configurations with 2, 4, 8, and 16 attention heads. A comparison between Variants I and II shows that increasing the number of attention heads from 2 to 4 results in significant performance gains across all datasets. However, further increasing the heads to 8 or 16 (Variants III and IV) causes a decline in detection accuracy. These findings indicate that a greater number of attention heads does not necessarily translate to better performance. In our proposed multimodal fusion module, setting the number of attention heads to 4 strikes the best balance and yields optimal results.

\begin{table}[]
\centering
\caption{Ablation study on the effect of attention mechanism in MFM} \label{taba}
\begin{tabular}{ccccc}
\hline
Setting       & FF++   & CDFv2  & DFDC   & Fsh    \\ \hline
w/o Attention & 0.9567 & 0.7587 & 0.6955 & \textbf{0.6162} \\
w/ Attention  & \textbf{0.9735 }& \textbf{0.7621} & \textbf{0.7003} & 0.6089 \\ \hline
\end{tabular}
\end{table}

\begin{table}[]
\centering
\caption{Ablation study on number of attention heads in MFM}\label{attn}
\begin{tabular}{cccccc}
\hline
Variation &Number & FF++            & CDFv2           & DFDC            & Fsh             \\ \hline
I &2       & 0.9704          & 0.7665          & 0.7028          & 0.6416          \\
II &4       & \textbf{0.9746} & \textbf{0.7751} & \textbf{0.7294} & \textbf{0.6575} \\
III &8       & 0.9736          & 0.7565          & 0.6928          & 0.6487          \\
IV &16      & 0.9732          & 0.7496          & 0.6908          & 0.6423          \\ \hline
\end{tabular}
\end{table}


In summary, the ablation studies confirm the significant contributions of the EfficientNet backbone, the pre-fusion module, and an appropriately configured number of attention heads to the overall model performance. The effective integration of these components substantially enhances detection accuracy across multiple datasets.

\subsubsection{T-SNE Visualization}

To illustrate the effectiveness of M3D-net in learning discriminative and generalizable representations, we conduct a t-SNE visualization of the learned feature distributions under cross-dataset settings. Specifically, we compare the features extracted from the standalone RGB branch with those from our full model. The visualization is conducted on unseen samples from the Celeb-DF-v1 dataset.

\begin{figure}[t]
    \centering
    \begin{subfigure}[t]{0.48\textwidth}
    \includegraphics[width=\textwidth]{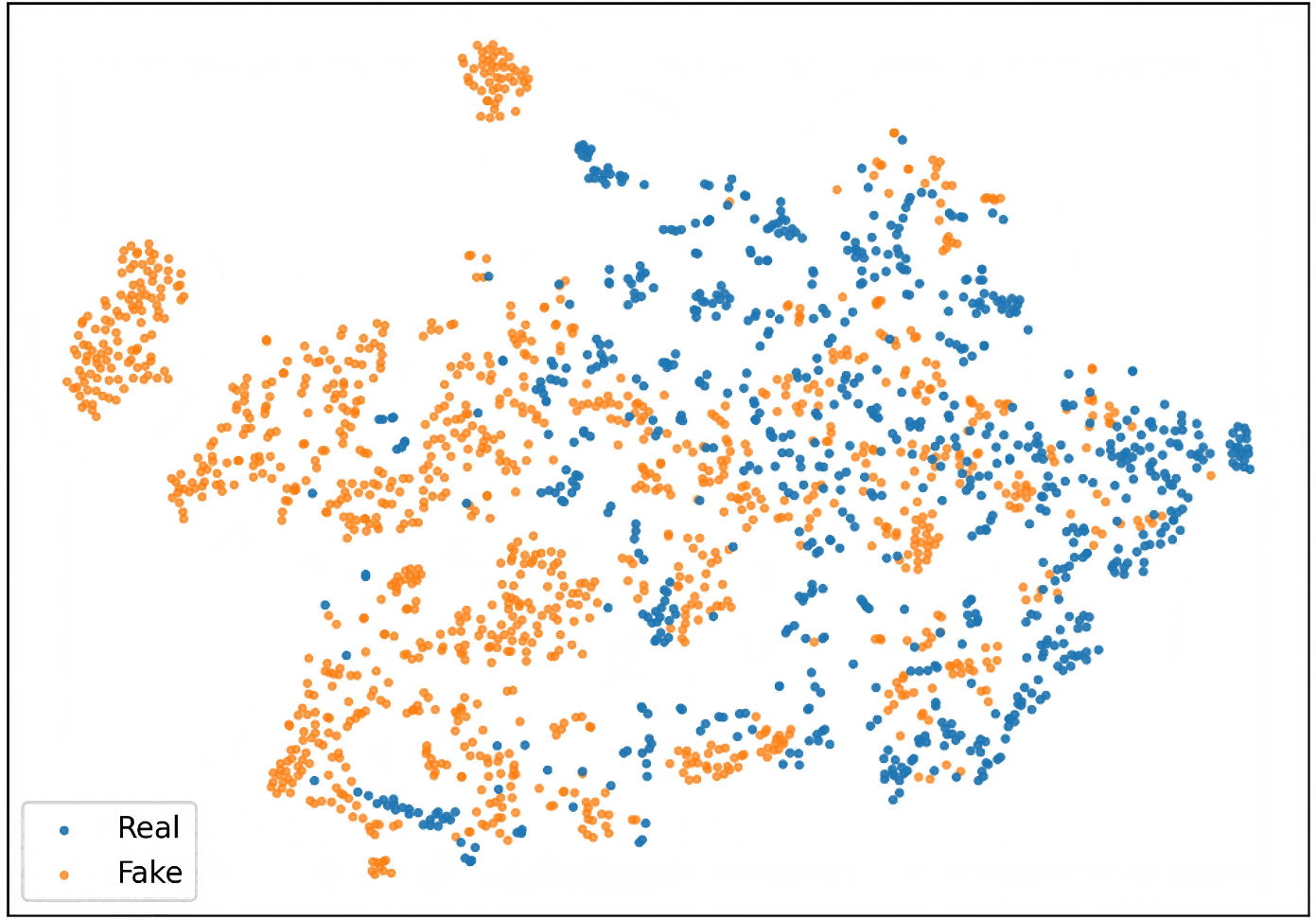}
    \caption{RGB branch}
    \label{fig:sub1}
  \end{subfigure}
  \hfill
  \begin{subfigure}[t]{0.48\textwidth}
    \includegraphics[width=\textwidth]{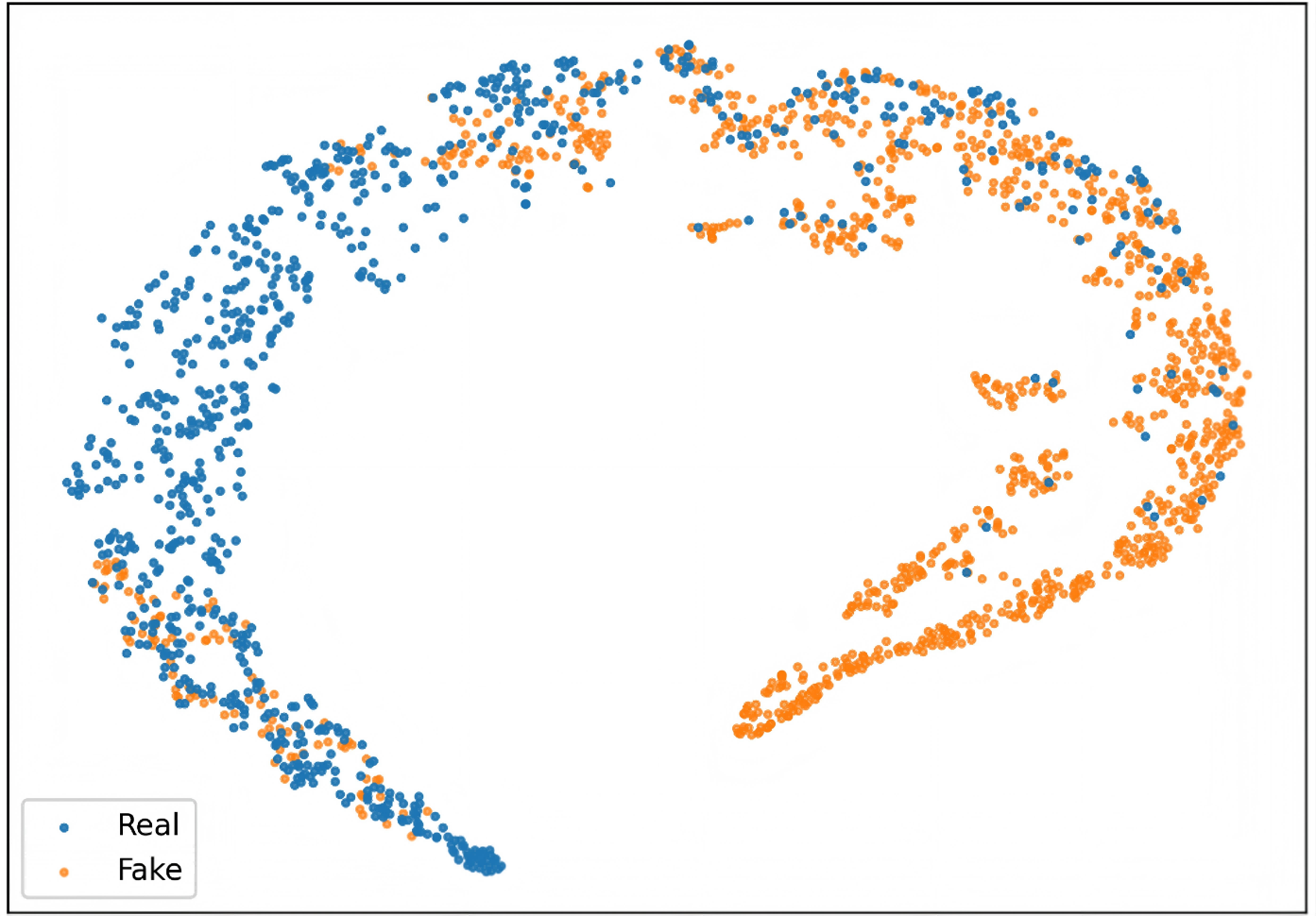}
    \caption{M3D-Net}
    \label{fig:sub2}
  \end{subfigure}
    \caption{t-SNE visualization of feature distributions on unseen samples from Celeb-DF-v1 under a cross-dataset setting. (a) Features from the standalone RGB branch show entangled real/fake clusters with poor separation. (b) Features from the full M3D-Net model exhibit well-separated and compact clusters, demonstrating enhanced discriminability and generalization through 3D facial reconstruction.}
    \label{tsne}
\end{figure}

As shown in Figure~\ref{tsne}, under the cross-dataset setting (Celeb-DF-v1), the RGB baseline cannot effectively distinguish between real and fake samples, as its t-SNE plot exhibits a highly entangled distribution with no clear separation between authentic and manipulated faces. This indicates that simply using RGB features is insufficient to extract meaningful representations for forgery detection. In contrast, our M3D-net forms well-separated and compact clusters in the feature space, achieving clear boundaries between real and fake samples. This demonstrates that the 3D facial reconstruction branch is capable of generating more discriminative and semantically structured feature representations.

\section{Conclusions}\label{sec5}
In this paper, we presented M3D-Net, a novel end-to-end deepfake detection framework that effectively leverages 3D facial feature reconstruction and multi-modal fusion. Our method introduces a self-supervised 3D reconstruction module that enables the decomposition of depth and albedo from single RGB images, thereby capturing subtle geometric and textural inconsistencies indicative of facial manipulation. To further enhance feature representation, we designed a 3D feature pre-fusion module (PFM) that adaptively recalibrates multi-scale features, and a cross-modal fusion module (MFM) that seamlessly integrates RGB and 3D-reconstructed features through attention-based interaction.
Comprehensive experiments on multiple public deepfake benchmark datasets validate that M3D-Net attains state-of-the-art (SOTA) detection accuracy and robustness.

The algorithm proposed in this paper still has certain limitations,\textit{ i.e.}, it relies on the symmetry of the target image. This enables the algorithm to achieve more favorable reconstruction performance for frontal symmetric faces, whereas its effectiveness degrades when dealing with faces in extreme poses or with occlusions.
In future work, we plan to further improve the framework by exploring advanced data augmentation strategies to enhance generalization, investigating more robust unsupervised 3D face reconstruction techniques that reduce reliance on symmetry priors, and extending the model to video-based deepfake detection by incorporating temporal consistency. 

\section*{Acknowledgements}
This work was supported by the Natural Science Foundation of Guangdong Province, China (2026A1515011792, 2026A1515010920); the Guangdong Provincial Key Laboratory, China (2023B1212060076).

\section*{Author contributions}
H.W. conceptualized the algorithm framework and took primary responsibility for manuscript drafting. Y.C. carried out the algorithm implementation and validation, and contributed to manuscript revision and editing. S.B., as the corresponding supervisor, provided overall research guidance, critically revised the manuscript, and ensured academic integrity. C.W. conducted independent manuscript review, offering valuable insights and constructive suggestions for content improvement.

\section*{Data availability statement}
No datasets were generated or analysed during the current study.

\section*{Conflict of interest}
The authors declare that they have no conflict of interest.

\bibliography{ref1}

\end{document}